\documentclass[letterpaper]{article}

\usepackage{natbib,alifeconf}
\usepackage{verbatim}
\usepackage{todonotes}
\usepackage[hidelinks]{hyperref}
\usepackage[shortlabels]{enumitem}
\usepackage{etoolbox}
\usepackage{subfig}

\newtoggle{ShowNames}
\toggletrue{ShowNames}
\setlist[itemize]{noitemsep}

\newcommand{\githubJHS}{\url{https://github.com/jstovold/ALIFE2023}}

\title{Neural Cellular Automata Can Respond to Signals}
\iftoggle{ShowNames}{
  \author{James Stovold \\ 
  \mbox{} \\ 
  School of Computing and Communications \\ 
  Lancaster University Leipzig \\
  Nikolaistra\ss{}e 10 \\ 
  Leipzig 04109, Germany \\ 
  Email: \url{j.stovold@lancaster.ac.uk}}%
}{ %
  \author{A. N. Author$^1$ \\ 
  \mbox{} \\ 
  $^1$Address Line 1 \\
  Address Line 2 \\
  Address Line 3 \\
  Address Line 4 \\
  Email: \url{a.n.author@institution.com} }
}

\begin{document}

 \maketitle

\begin{abstract}

%  Signals are an essential part of animal life. 
 
  Neural Cellular Automata (NCAs) are a model of morphogenesis, capable of growing two-dimensional artificial 
  organisms from a single seed cell. In this paper, we show that NCAs can be trained to respond to signals. Two types 
  of signal are used: internal (genomically-coded) signals, and external (environmental) signals. Signals are 
  presented to a single pixel for a single timestep. 
  
  Results show NCAs are able to grow into multiple distinct forms based on internal signals, and are able to change 
  colour based on external signals. Overall these contribute to the development of NCAs as a model of artificial 
  morphogenesis, and pave the way for future developments embedding dynamic behaviour into the NCA model.

 Code and target images are available through GitHub: \githubJHS{}

\end{abstract}

% interested in whether similar behaviour can be included within the models.

\section{Introduction}

Signals are the central component of animal interaction, and an essential part of animal 
life~\citep{maynardsmith_animalsignalsmodels}.  We can distinguish between internal signals (signals within the body 
itself), societal signals (external to the body but internal to the social group of the animal), and external signals 
(from the environment or from other animals).

Internal signals include hormonal or neural signals within the body~\citep{widmaier_vandershumanphysiology}, 
bioelectrical~\citep{adams_endogenousvoltagegradients} and genomic signals during 
development~\citep{jaenisch_epigeneticregulationgene,wang_genomewidestudy}. External signals include the basic senses 
and detecting gradients~\citep{parent_cellssensedirection}. Societal signals include the famous waggle 
dance~\citep{lindauer_communicationswarmbees, camazine_househuntinghoney} and its inhibitory 
headbutt~\citep{seeley_stopsignalsprovide}, the pheromones released by foraging 
ants~\citep{deneubourg_blindleadingtheblind}, identity formation signals~\citep{stovold_preservingswarmidentity}, 
aggregation signals of {\em Dictyostelium discoideum}~\citep{gross_signalemissionsignal}, dog 
sneezes~\citep{walker_sneezetoleave}, and---of course---human language and communication. Understanding and studying 
signals is clearly an important topic; as researchers in artificial life, it is reasonable to ask how this can be 
included in our models.

\citet{mordvintsev_growingneuralcellular} developed the `Neural Cellular Automata' (NCA) as a model of morphogenesis, 
to study the growth of an artificial organism from a single cell. By exploiting the power of artificial neural 
networks to derive the rules of interaction in a cellular automata~\citep{wolfram_cellularautomatamodels, 
adamatzky_gameoflife} the NCA is able to produce two-dimensional artificial organisms. What NCAs lack, however, is a 
mechanism for responding to signals.

%without the ability to interact with its environment or other animals, no model of an artificial organism will be 
%complete.

This paper presents an approach to training NCAs such that they are able to respond to signals. Two types of signal 
are provided to the NCA: internal and external. For internal signals, we encode the signal into the seed cell, showing 
that the NCA can grow into different shapes and colours according to the signals provided. For external signals, we 
provide the signal after the NCA organism has fully grown, and show that the organism can consistently and repeatedly 
respond to the external signal to change its colour.

Previous approaches to including signal responses in NCAs consist of moving the entire organism up an environmental 
gradient~\citep{kuriyama_gradientclimbingneural}, and replacing the organism with a new organism that has a different 
property (say, colour) by introducing a new seed cell~\citep{cavuoti_adversarialtakeoverneural}. In contrast to these 
approaches, the work presented in this paper uses a single NCA to respond to a signal that is only presented through a 
single cell for a single timestep. This means that the grown organism is \emph{responding} to a signal rather than the 
signal acting as a new seed cell triggering the growth of a distinct organism.

%treating the signal as a new seed cell for a distinct organism.

\section{Neural Cellular Automata} 
\begin{figure*}[t!]
 \centering
 \includegraphics[width=\linewidth]{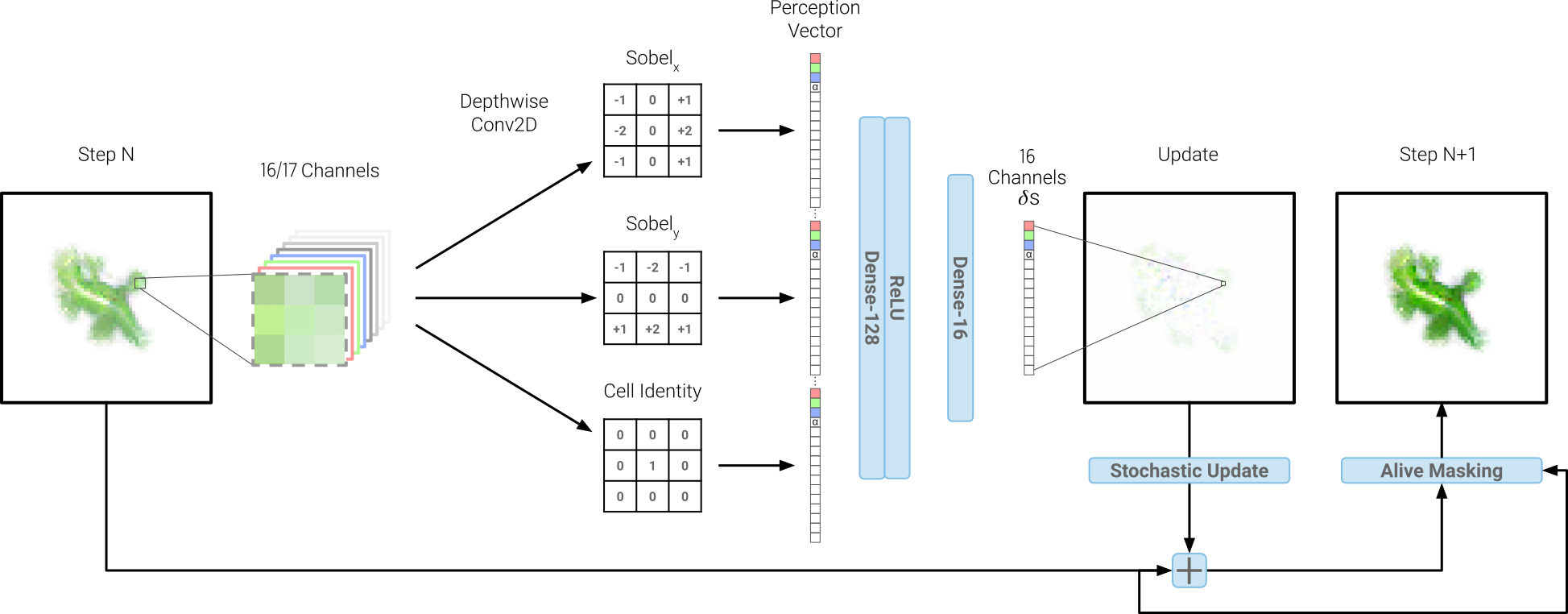}
 \caption{Diagram depicting one pass of the NCA update step. The diagram also shows the structure of the neural 
 network. Image adapted from \citep{mordvintsev_growingneuralcellular}, licenced under CC BY 4.0.}
 \label{fig:nn:structure}
\end{figure*}

A conventional Cellular Automata (CA) consists of a two-dimensional grid of cells. Each cell can be in one of two 
states: alive or dead. Each cell contains a single automaton which updates the state of the cell based on the state of 
its neighbouring cells. The rules of the automaton were originally designed by hand~\citep{izhikevich_gameoflife}, but 
a variety of techniques (such as genetic algorithms) have also been employed to find CAs with various interesting 
characteristics~\citep{mitchell_evolvingcellularautomata}. Multiple variants of CAs have been developed, with many 
focussing on their application to biological 
phenomena~\citep{ermentrout_cellularautomataapproaches,alber_oncellularautomaton,chan_leniabiologyartificial},

Neural Cellular Automata (NCAs)~\citep{mordvintsev_growingneuralcellular} extend the conventional 2D CA so that the 
state of each cell is now a vector of 16 real-valued numbers rather than just alive/dead and replaces the update rule 
with an artificial neural network. The artificial neural network is trained to update the state of each cell in such a 
way that a certain macroscopic form will emerge from a single seed cell.

\begin{figure}[h!]
 \centering
 \includegraphics[width=0.75\linewidth]{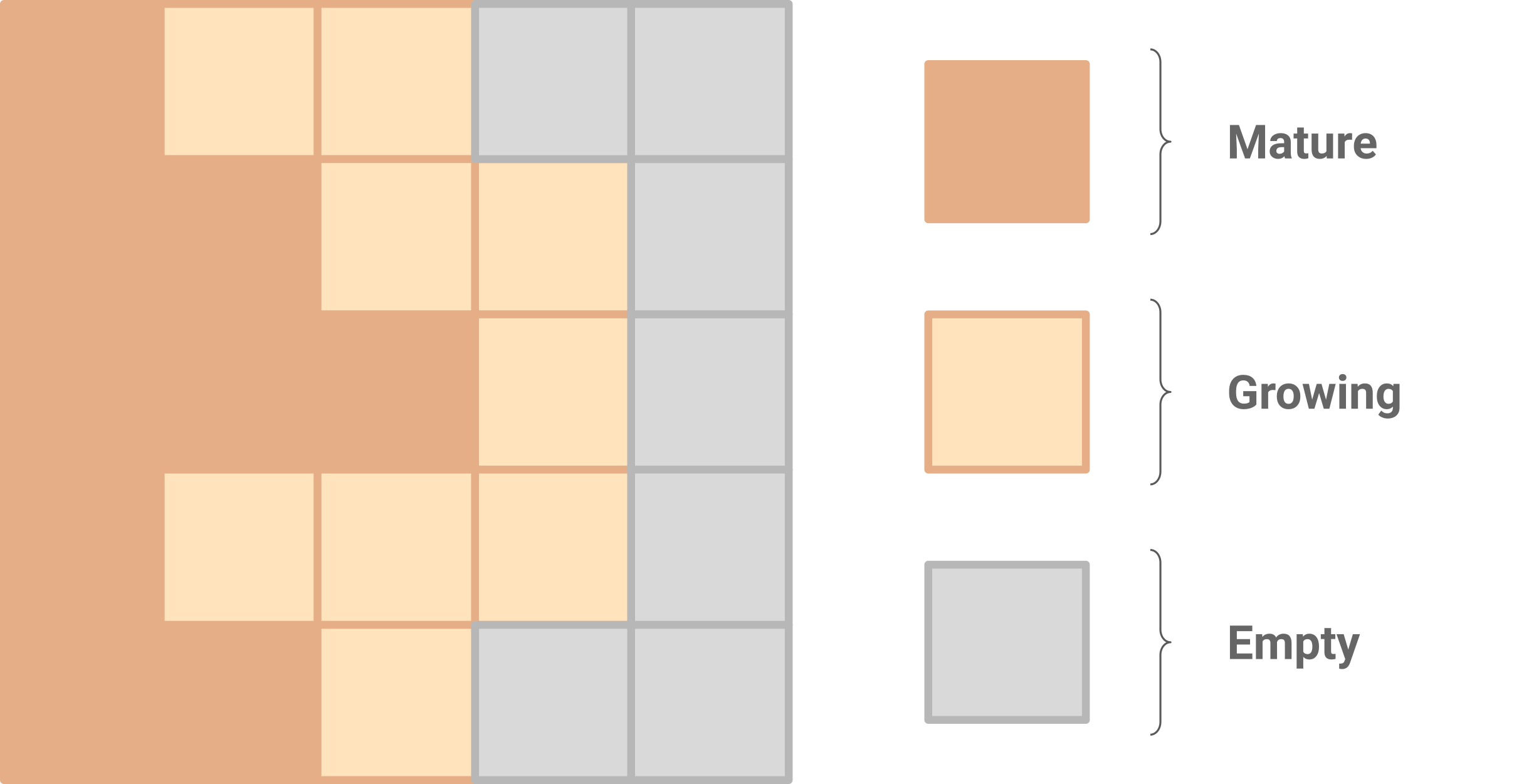}
 \caption[]{Diagram depicting the role of the alpha channel in the NCA for indicating which cells are alive and 
 which are eligible to come to life in the next step. Image from \cite{mordvintsev_growingneuralcellular}, licenced 
 under CC BY 4.0.}
 \label{fig:alpha}
\end{figure}

In previous work on NCAs, the state vector in each cell consists of the same form: three visible channels representing 
red, green, and blue of each pixel in the organism, an alpha channel which represents the maturity stage of the cell 
(see fig.~\ref{fig:alpha}), and 12 `hidden' channels which cells use to communicate. The seed cell for a new organism 
is then a single cell with all values set to 1, except for the RGB channels, which are set to 0. There are 
occasionally alternative approaches employed, such as multiple seed cells to orient the organism in 
space~\citep{mordvintsev_growingisotropicneural}; or multiple alpha channels to combine two NCAs into 
one~\citep{cavuoti_adversarialtakeoverneural}.

Training an NCA consists of running the CA from a seed cell for a set number of steps (typically in the range 
64--200), then comparing the grown organism to a target image. The comparison is used to produce a loss signal which 
is fed back to the neural network for training via the backpropagation 
algorithm~\citep{rumelhart_learningrepresentationsbackpropagating}. Figure \ref{fig:nn:structure} depicts one forward 
step of the NCA, showing how the neural network acts as the update rule for the CA.

\citet{mordvintsev_growingneuralcellular} introduce three approaches to training the NCA, labelled `growing', 
`persistent', and `regenerating'. While all three use a batch training approach, the manner in which the batch is 
manipulated differs between them. In `growing', the NCA is trained as described above, with the CA iterated for a set 
number of steps, then the organism compared with a target image.

\begin{figure}[h!]
 \centering
\hrule
 \includegraphics[width=\linewidth]{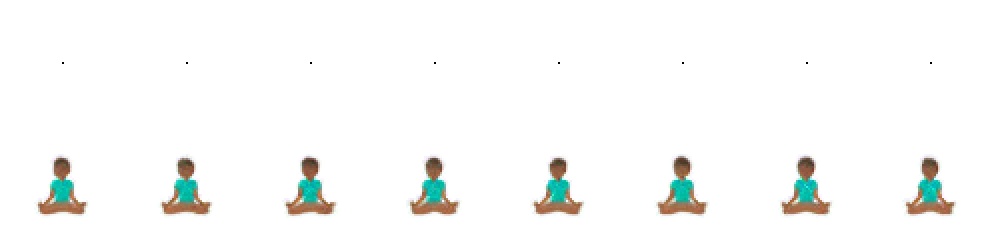} \hrule %[2ex] 
 \includegraphics[width=\linewidth]{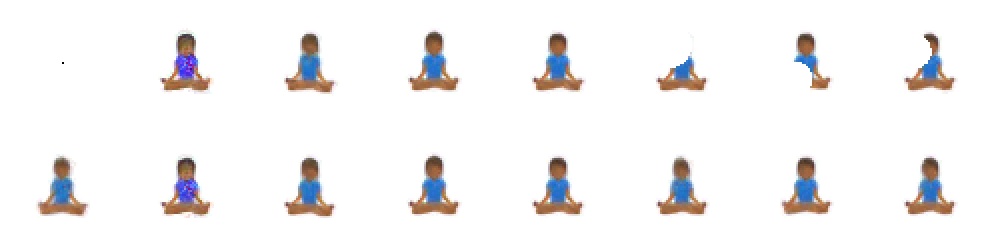}
\hrule 
\vspace{1em}

\caption[]{Example batches from the `growing' (top), and `regenerating' (bottom) training regimes. Each example 
comprises two rows, the upper row shows the initial state of the CA, the lower row shows the state of the CA after it 
has been iterated for between 64 and 200 steps (the actual number of steps is randomly picked each time). The damage 
caused to the organism is clearly seen in the three rightmost NCAs of the regenerating example. Horizontal lines added 
to help distinguish the two examples. }

 \label{fig:batch}
\end{figure}

In `persistent' and `regenerating', however, the CA is grown from a seed, but once it is grown it is placed back into 
a training pool. At each training step, the batch is drawn from the training pool by random sample, meaning some 
fully-grown organisms are included in the batch (see fig.~\ref{fig:batch}). These fully-grown organisms are iterated 
in the same way as the original CA, training the NCA to produce organisms that are stable over extended periods of 
time.

For the `regenerating' training regime, the same batch approach is used as with `persistent', but a small portion of 
the grown organisms extracted from the training pool are damaged before they are iterated. This damage introduces a new 
signal to the neural network that it should learn how to maintain the organisms even in the presence of external 
interference. The `regenerating' approach to training NCAs forms the basis of our external signal training approach 
(detailed below).

\section{Methods}

In this paper, we present two types of signal to the NCA: {\em internal signals}, which are provided through the seed 
cell (similar to genomic signals during development); and {\em external signals}, which are provided after the 
organism is fully developed.

Internal signals are presented to the NCA without requiring any structural changes to be made. For the external 
signals, however, we extend the state vector to include a read-only `environment' channel. This channel allows us to 
provide a signal to a specific cell by setting that channel to 1 for a single timestep. As this increases the number 
of channels which the NCA has on its input (but not its output), we end up with a slightly different structure to the 
neural network.

Fig.~\ref{fig:nn:structure} shows the overall structure of a single update step for the NCA. This is mostly the same as 
for the original NCA update step defined by \citet{mordvintsev_growingneuralcellular} except that we have 17 input 
channels and a `perception vector' of size 51, rather than 16 inputs and a perception vector of 48. The number of 
output channels remains the same at 16, the 17th channel is read-only so the NCA is not able to make changes, only to 
receive signals from it.

\begin{figure}[h!]
 \centering
 \includegraphics[width=\linewidth]{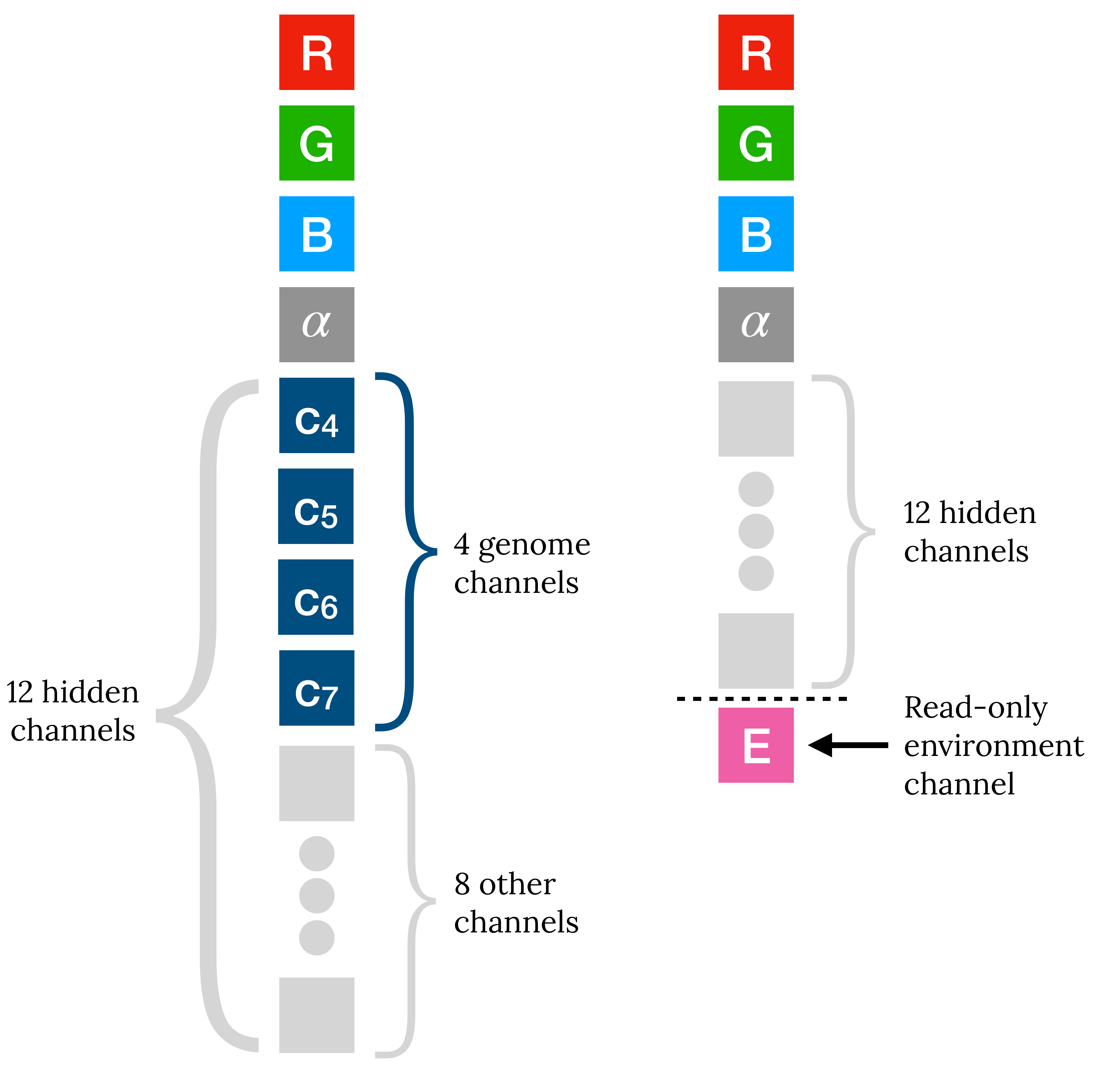}
 \caption[]{Structure of the NCA seed cell (a) for internal signals, and (b) for reacting to external signals. In (a), 
 we have 4 genome channels, but this could be increased or decreased by adjusting the number of remaining hidden 
 channels accordingly. In no experiments were more than 12 hidden channels used.} 
 \label{fig:seed}
\end{figure}

Fig.~\ref{fig:seed} shows the structure of our updated seed cells. In both cases the first four channels are the same 
as other implementations of NCAs, in that $c_0$--$c_2$ code for the current RGB colour value of the cell, and the alpha 
channel ($c_3$) codes for the maturity of the cell. 

When considering internal signals, we encode information into the seed cell to signal to the NCA which organism it 
should grow. The following $n$ channels ($c_4$--$c_{4+n-1}$) encode this length-$n$ binary `genome'. The remaining 
channels are set to 1.0 in the seed cell and represent the space available for cells to communicate with one another. 
It is worth highlighting that, while the seed cell has the genome encoded into the $n$ channels $c_4$--$c_{4+n-1}$, 
this is only provided to the seed cell, and it is up to the NCA to remember and use that information---it is not 
provided to every cell at every time step.

For external signals, we add an extra channel to the end of the cell state vector which the neural network can read 
from, but not write to. This allows us to introduce signals to the organism as if from the environment. For the present 
paper, this channel remains at 0 until we introduce a signal at which point a single cell will have the environment 
channel set to 1 for a single time step.

\subsection{Training Approach}

All the models presented in this paper were trained from a single seed cell, with a single neural network trained for 
each experiment (rather than training a different network for each organism). 

\subsection{Internal Signals}

To train the NCA to respond to internal signals, we provide different target images depending on the genomic sequence 
in the seed cell (as depicted in fig.~\ref{fig:seed}). As the signal is only provided at the start, the NCA needs to 
learn how to remember that signal as it grows, because it will only be evaluated against the target organism after it 
has finished growing (64--200 steps later). The error between the target and the grown organism is then used to train 
the neural network via backpropagation.

\begin{figure}[h!]
 \centering
 \includegraphics[width=\linewidth]{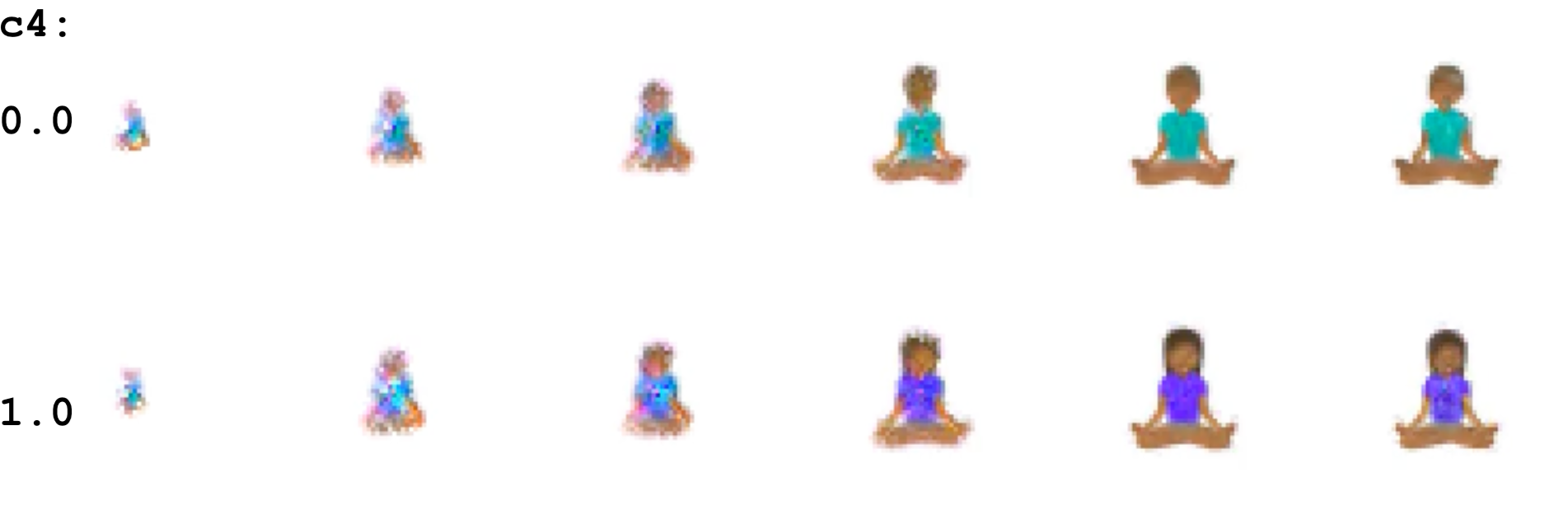}\\
 \includegraphics[width=\linewidth]{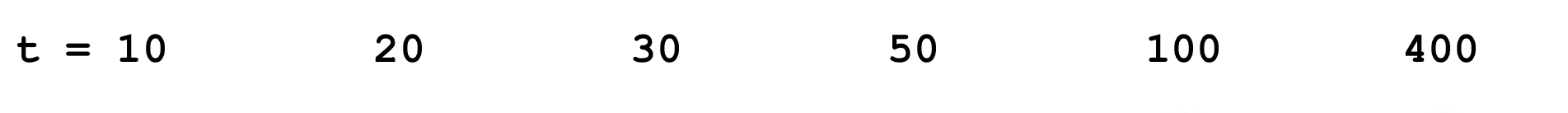}

 \caption[]{Growth of two meditation emoji from a single NCA trained to produce one of two organisms depending on a 
 single change in the seed cell. The only difference between the two is $c_4 = 0$ for the top example, $c_4 = 1$ for 
 the bottom example. The NCA is trained to produce the organism within 200 timesteps. Snapshots are shown from times 
 $t = 10, 20, 30, 50, 100, 400$.}
 \label{fig:param:meditate}
\end{figure}

An early result is shown in fig.~\ref{fig:param:meditate}, where we used a single genomic bit ($c_4$) to distinguish 
between growing two distinct forms of meditation emoji. The signal is only provided to the NCA at the start of the run 
(i.e.\ in the seed cell).\footnote{interestingly, a separate experiment was run to try and encourage the NCA to keep a 
copy of the genome within each cell, similar to `real' cells and their DNA, but while this trained more quickly, it 
appeared to produce less stable organisms}

\subsection{External Signals}

When training the NCA to respond to external signals, we had to adjust the training regime so that the new signal 
could be introduced during training. The approach taken was to co-opt the concept of damaging the organism from the 
`regenerating' regime to introduce an external signal instead. Furthermore, we kept track of the number of times we 
provided the signal to the organism, to push the NCA to learn how to switch back and forth between two similar 
organisms. This was in direct response to an early attempt which saw the NCA learn to change in one direction, but 
when it changed back again it looked similar but had lost the ability to respond to signals (meaning it was actually a 
different organism). The other benefit of co-opting the damage part of training was that it prevented the signal being 
inadvertently lost by providing the signal at the same time as the damage.

A batch of size 12 was used for training the signal response: the 12 organisms were sorted based on their current loss, 
with the 3 worst performing being removed from the batch and replaced by a new seed cell. The rest of the batch were 
split into (a) those which should `persist' (i.e.\ they were not provided a signal, but had to continue to exist for 
another round, preventing the organism from degrading) and (b) those which were provided a signal, and so had to change 
some property (typically their colour). Organisms that receive the external signal had target images changed at the 
same time in order for them to learn the response.

\section{Results}

This section presents the models that were trained, and shows how the organism varies based on the signal provided. 
Internal signals, provided through the seed cell, are used in the first two tests to indicate which organism should be 
grown by the NCA; external signals, provided to the grown organism through the environment, are used in the third test 
to indicate to the organism that it should change colour.

\begin{figure}[h!]
 \centering
 \includegraphics[width=\linewidth]{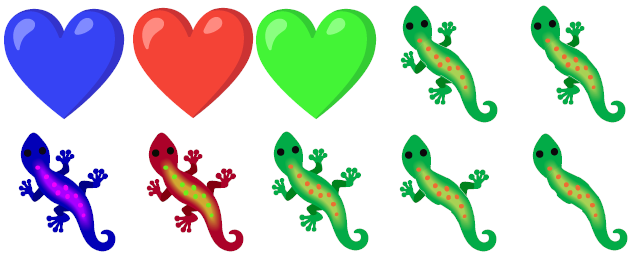}
 \caption[]{A subset of the target images used for training the NCAs. There are a further 11 geckos with different 
 missing legs that are not shown. }
 \label{fig:target_images}
\end{figure}

A subset of the target images used for training the NCAs are provided in fig.~\ref{fig:target_images} (there are 16 
variants of the green gecko with different limbs removed, only 5 are shown here).

\subsection{Internal Signals I: Different Forms}

This test aimed to see whether a single NCA was able to grow into different colours and shapes based on information 
encoded into the seed cell---our internal signals. Using the `growing' training approach, we tested two scenarios: (i) 
growing the same shape (a heart) in different orientations and different sizes, and (ii) growing two shapes (heart and 
gecko) in three colours (red, green, and blue). % using four channels of the seed cell.

\begin{figure}[h!]
 \centering
 \subfloat[Growth of heart shape with different sizes. 
   \label{fig:results:heart:size}] { \centering % 
\includegraphics[width=0.45\linewidth]{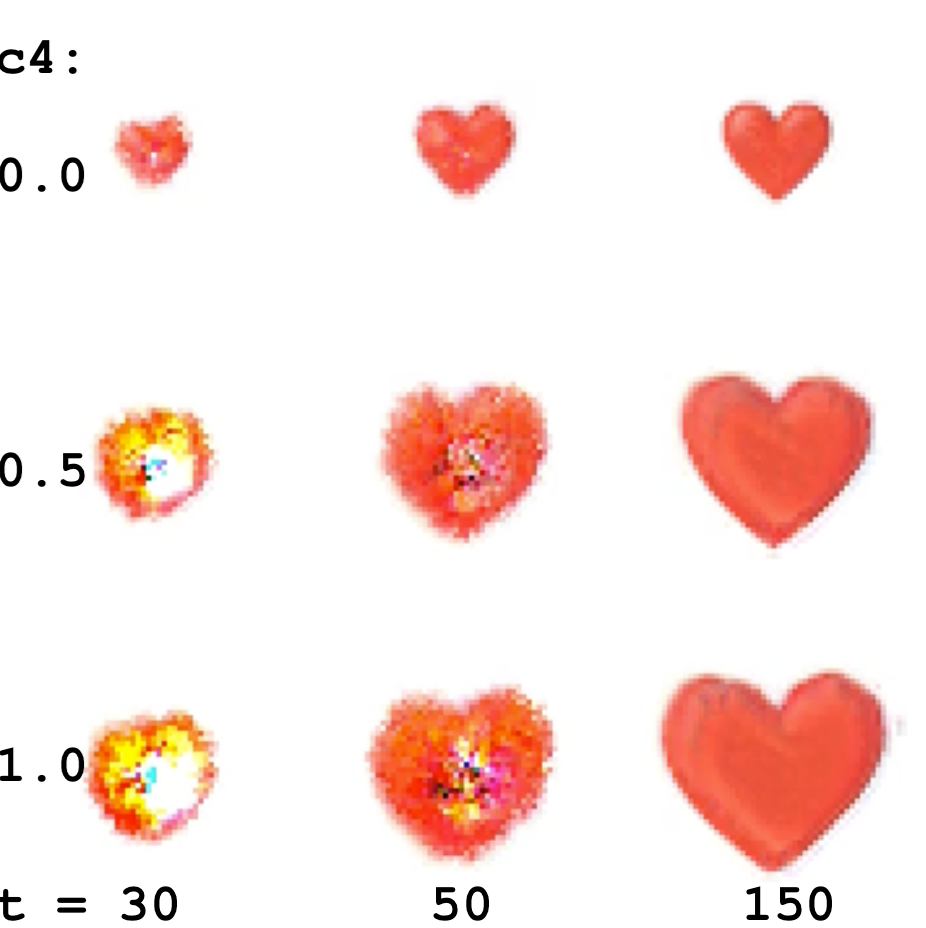} } \hfill{}
 \subfloat[Growth of heart shape with different orientations.  \label{fig:results:heart:orientation}] { \centering %
 \includegraphics[width=0.45\linewidth]{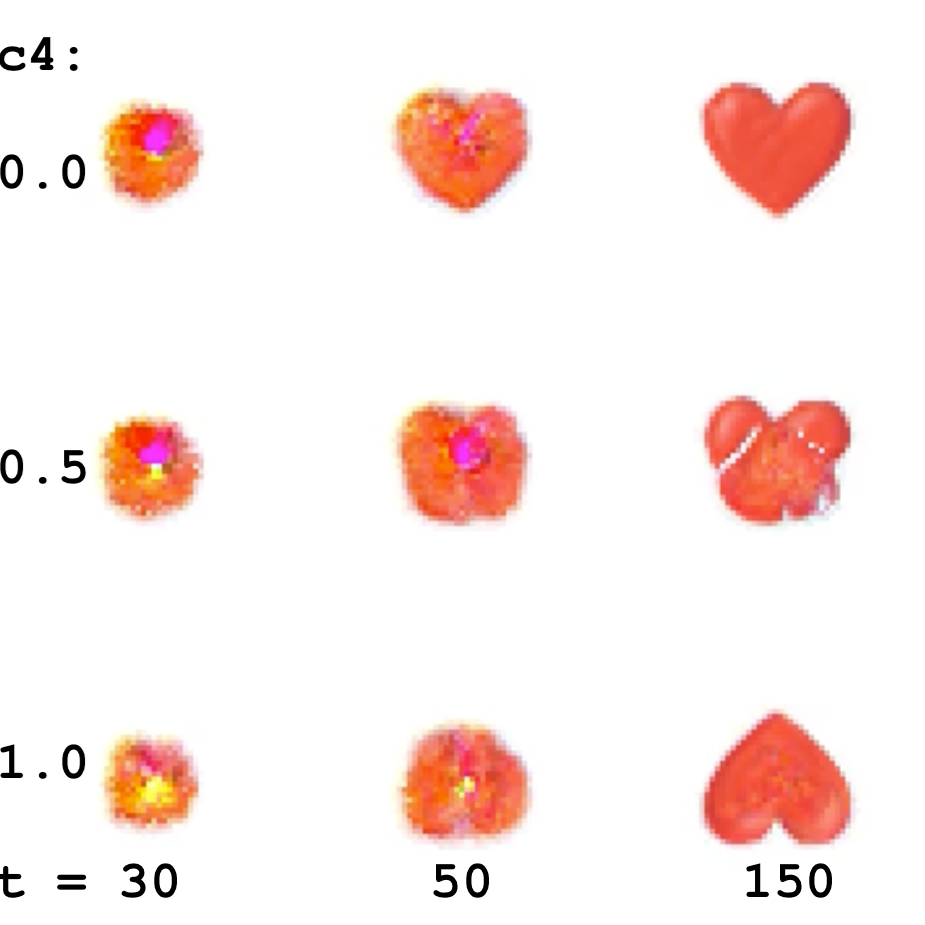} }
 \caption[]{In both cases, from left-to-right: $t=30,50,150$; from top to bottom: $c_4=0.0,0.5,1.0$. Only $c_4=0.0,1.0$ 
 were trained into the NCA, meaning $c_4=0.5$ is an out-of-training test in both cases.}

 \label{fig:results:heart}
\end{figure}

Figure~\ref{fig:results:heart} shows the outcome of scenario (i). For fig.~\ref{fig:results:heart:size} we trained the 
NCA to produce a small heart if we set $c_4$ to 0.0 and a large heart if we set $c_4$ to 1.0. 
Fig.~\ref{fig:results:heart:size} shows the growth of the NCA at $t=30, 50, 150$. The top row is when $c_4=0.0$, the 
bottom is $c_4=1.0$, and the middle is $c_4=0.5$. It is clear that the NCA has learnt to associate the size of the 
organism with that one internal signal (it is worth emphasising that at no point was the NCA shown the middle-sized 
heart).

For fig.~\ref{fig:results:heart:orientation} we similarly trained the NCA to produce an upright heart when $c_4 = 0.0$ 
and a rotated heart if $c_4 = 1.0$. Unlike the size example, when we give the NCA a signal of $c_4 = 0.5$ it is unable 
to produce something half-way between the two. This is not unexpected, given the heart was rotated by 180\textdegree{} 
but we also tested 90\textdegree{} and 270\textdegree{} and found similar behaviour.

For scenario (ii), we were interested in pushing the limits of the neural network's capacity---how many distinct 
organisms could be trained into a single NCA? We trained the NCA to grow six organisms: red, green, and blue for both 
geckos and hearts (see leftmost 6 target images in fig.~\ref{fig:target_images}). To encode this into the seed cell, 
we used four of the hidden channels: $c_4$--$c_7$. Table~\ref{tab:results:multi_genome} shows the mapping of channel 
to phenotypic trait.

\begin{table}[h!]
 \centering
 \begin{tabular}{c|l}
  Channel  	&  Trait 	   		\\ \hline 
  $c_4$		&  shape (gecko or heart)  	\\ 
  $c_5$		&  red  			\\ 
  $c_6$		&  green		    	\\ 
  $c_7$		&  blue 		  	\\ 
 \end{tabular}

 \caption[]{Genome mapping to encode phenotypic traits into the NCA seed cell. For example, a genome of $(0,0,1,0)$ would 
 encode a green gecko and $(1,1,0,0)$ would encode a red heart. }

 \label{tab:results:multi_genome}
\end{table}

\begin{figure}[h!]
 \centering
 \includegraphics[width=.85\linewidth]{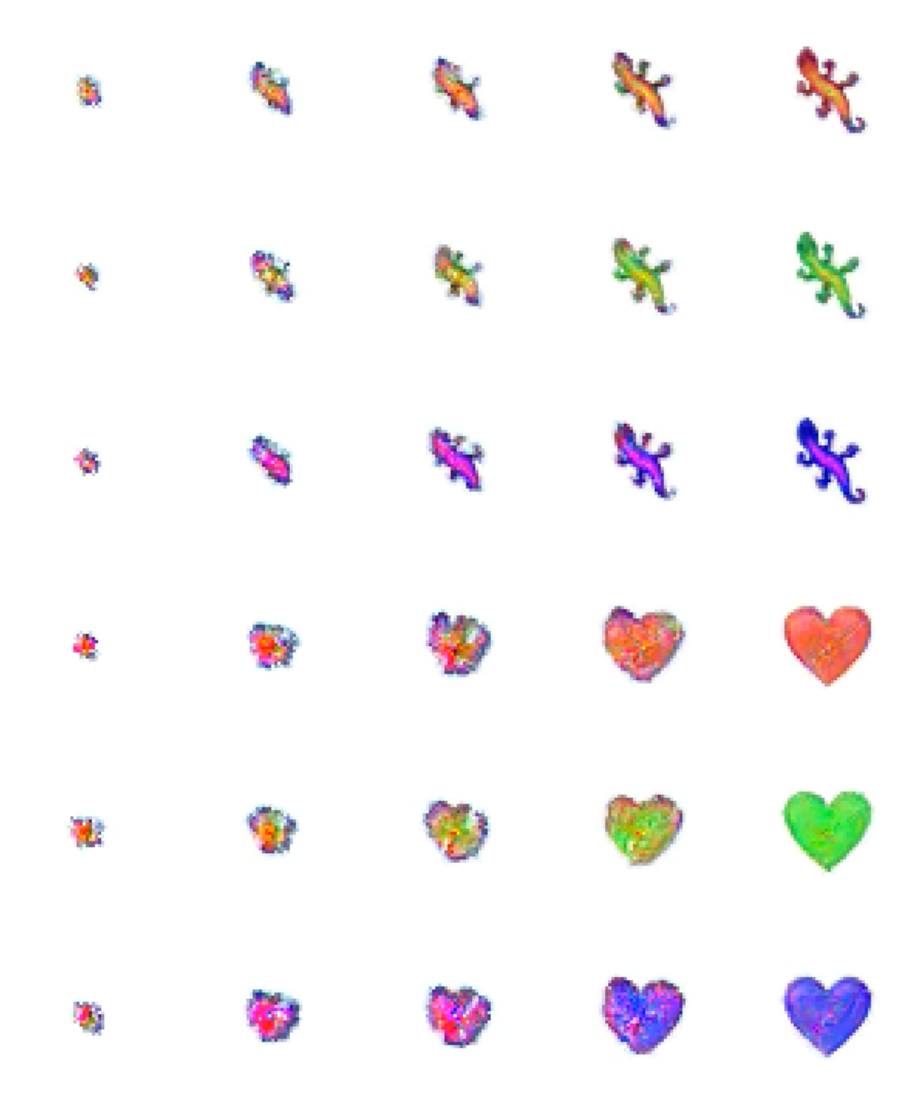} \\
 \includegraphics[width=.85\linewidth]{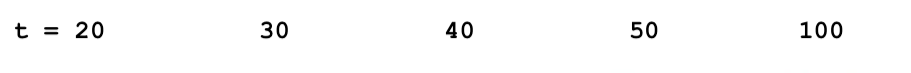}
 \caption[]{Growth of NCA trained with six distinct figures (three hearts and geckos of different colours); from left 
 to right: $t=20,30,40,50,100$ timesteps.}
 \label{fig:results:multi}
\end{figure}

\begin{figure}[h!]
 \centering
 \includegraphics[width=.85\linewidth]{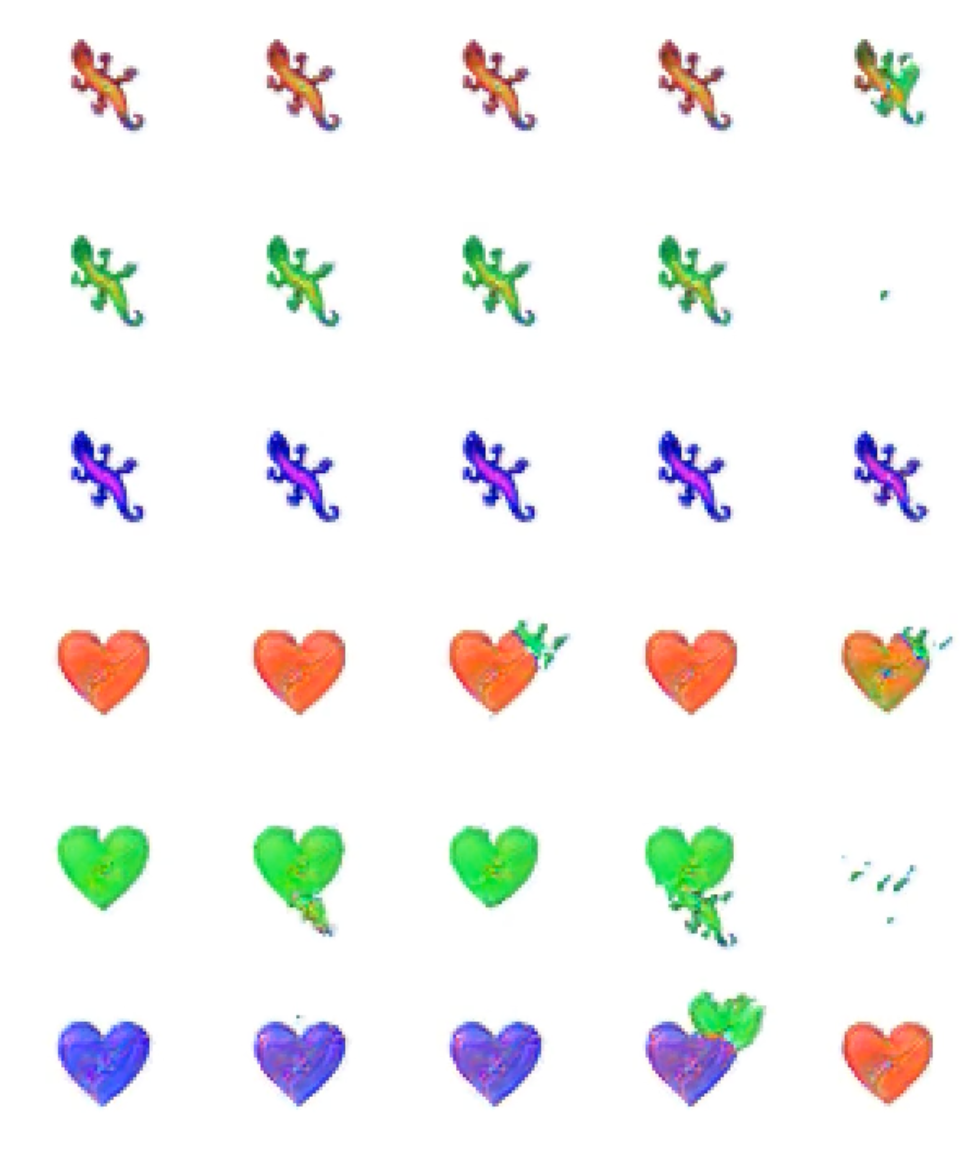} \\ 
 \includegraphics[width=.85\linewidth]{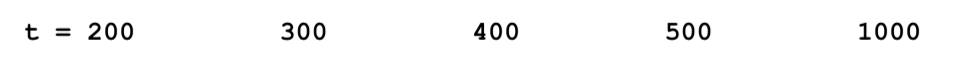}
 \caption[]{Extended stability test for multiple figures stored in NCA. The five columns represent the state of the NCA 
 after $t=200, 300, 400, 500, 1000$ timesteps.}
 \label{fig:results:multi_stability}
\end{figure}

Fig.~\ref{fig:results:multi} shows the outcome of this test. The NCA successfully learnt to grow all six different 
shapes with different colours. The NCA was trained to exist for up to 200 timesteps; 
fig.~\ref{fig:results:multi_stability} shows that past this point, the organisms have a tendency to either 
disintegrate, sprout other organisms that are trained into the network, or switch to another organism entirely (e.g.\ 
the blue heart switching to red after sprouting a green heart on the bottom row).

\begin{figure}[h!]
 \centering
 \includegraphics[width=\linewidth]{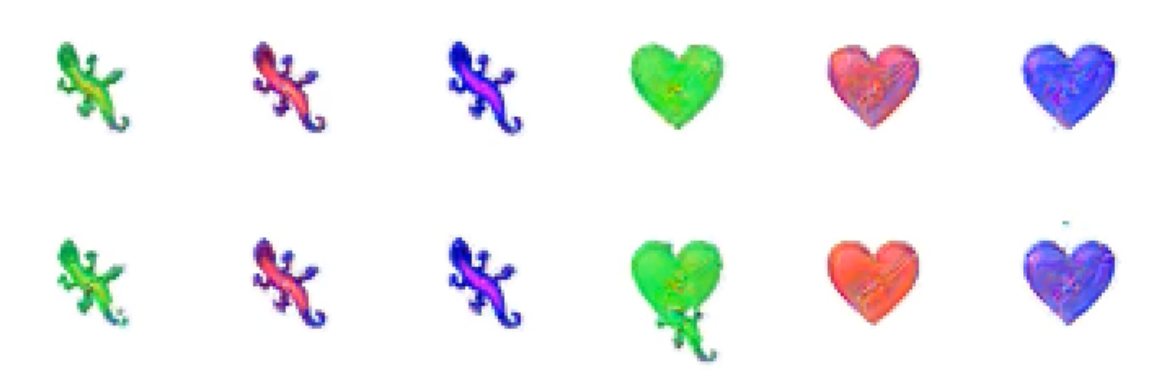} \\
 \includegraphics[width=\linewidth]{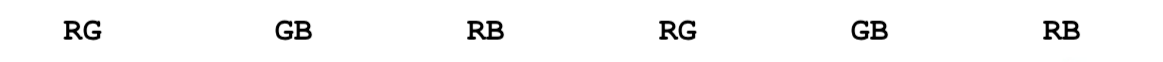}
 
\caption[]{Out of training colour test. From left to right: geckos with: red and green (RG), green and blue (GB), red 
 and blue (RB); similar for hearts; top is $t = 100$, bottom is $t = 400$.}
 \label{fig:results:multi_colour}
\end{figure}

Similar to the out-of-training tests we performed on the heart scenario above, we tested the ability of the NCA to 
generalise to other colours by providing seeds with multiple colour genes set to 1. 
Fig.~\ref{fig:results:multi_colour} shows the outcome of this test. While some of the organisms produced were part way 
between two colours---especially the purple gecko in the second column---closer inspection shows that this was not due 
to the information provided through the seed cell. In this particular example, the seed included green and blue, 
rather than red and blue as would have been expected for a purple gecko. Furthermore, it appears that the NCA also 
produces less-stable organisms when we include multiple colours in the seed cell. It is clear from these images 
that---with the current approach to training---the NCA is unable to generalise to other colours.

\subsection{Internal Signals II: Gecko Legs}

We wanted to study how the NCA remembers the signal as it grows. One possible mechanism is that it does not remember 
the signal itself, but that the properties of the organism are sufficiently different for each target that once it 
starts growing, the growing organism itself embodies the signal. For example, if we code for a blue gecko, once there 
are blue cells in the NCA, we know that it is either going to be a blue gecko or a blue heart.

%one possible reason why the previous scenario worked so well was the diversity of organisms we trained into the NCA.

To better understand how the NCA remembers the internal signal, we trained lots of very similar target organisms into 
the same NCA. Specifically, we designed the target organisms in such a way that one organism could easily be changed 
into another. The target images were all green geckos, but each with a different arrangement of legs (see rightmost 
four images of fig.~\ref{fig:target_images}).

By restricting the difference between geckos to just the legs, the core body shape, head, and tail were all present in 
all targets. We then use the damage process of the `regenerating' training regime (whereby we remove a chunk of the 
image to see how well it can grow back) to remove one of the legs, effectively changing the image into one of the 
other targets. If the NCA is able to re-grow the correct legs, then we know that it is able to remember which organism 
is being grown based on the information provided in the seed cell rather than relying solely on the existing shape of 
the grown organism.

In this test, we encode 16 variants of the gecko into four channels in the seed cell, where each gene signals whether 
a particular leg $l$ should be grown ($c_l = 1$) or should not ($c_l = 0$); see table~\ref{tab:results:leg_genome}.

\begin{table}[h!]
 \centering
 \begin{tabular}{c|l}
  Channel  	&  Trait 	   	\\ \hline 
  $c_4$		&  front-left leg  	\\ 
  $c_5$		&  front-right leg  	\\ 
  $c_6$		&  back-left leg    	\\ 
  $c_7$		&  back-right leg  	\\ 
 \end{tabular}
 \caption[]{Genome mapping to encode phenotypic traits into the NCA seed cell. For example, $(1,1,1,1)$ encodes for all 
 four legs.}
 \label{tab:results:leg_genome}
\end{table}

\begin{figure}[h!] 
 \centering
 \includegraphics[width=\linewidth]{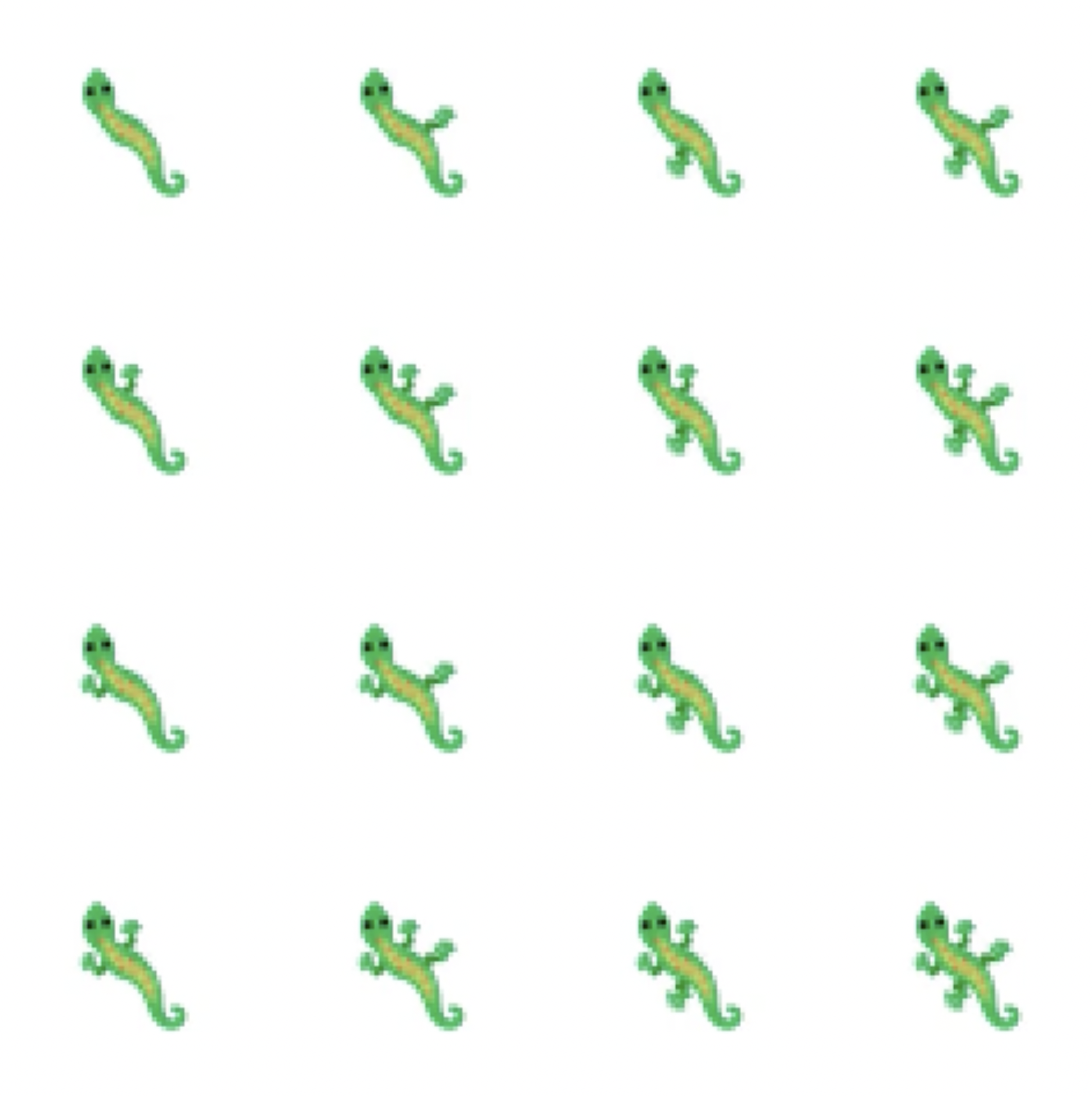}
 \caption[]{Fully-grown form ($t=200$) of all sixteen distinct geckos from a single NCA, trained using the genomic 
 mapping given in table~\ref{tab:results:leg_genome}. }
 \label{fig:gecko_legs}
\end{figure}

Fig.~\ref{fig:gecko_legs} shows the result of training the NCA. Each organism is grown from a distinct seed cell, 
according to the mapping given in table~\ref{tab:results:leg_genome}. All sixteen variants of the gecko are trained 
into the same NCA, demonstrating that an NCA is able to grow many similar organisms as well as the highly-distinct 
organisms shown before.

\begin{figure}[h!]
 \centering
 \includegraphics[width=\linewidth]{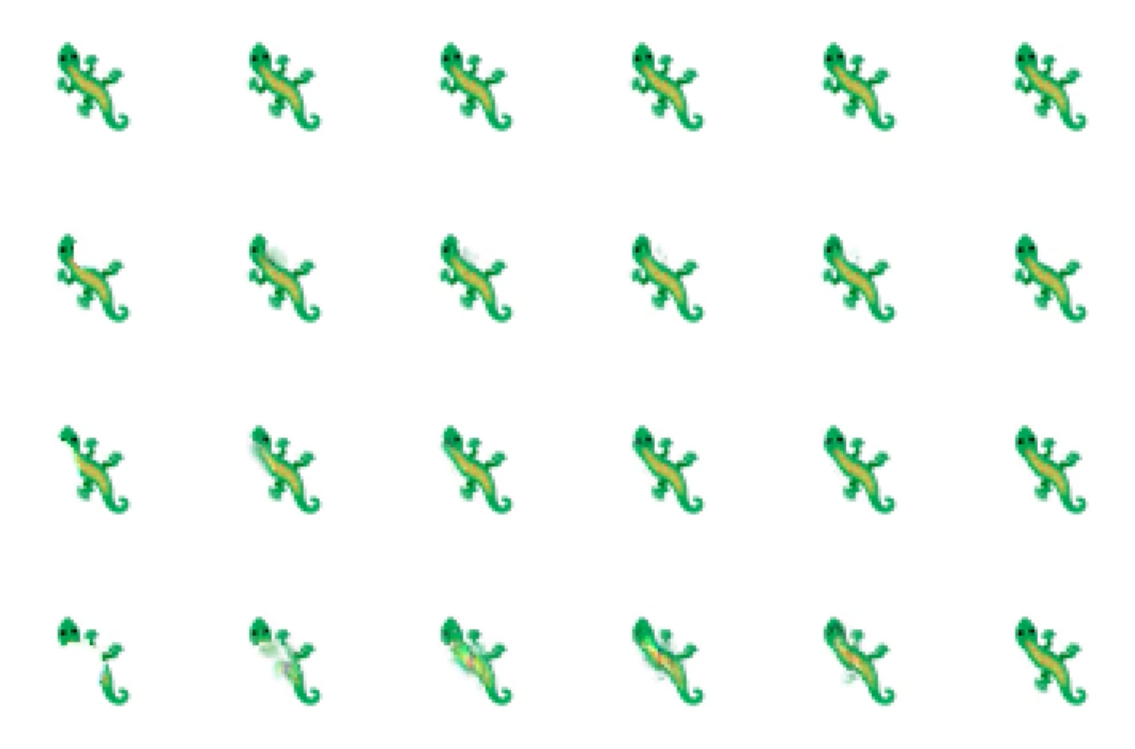} \\
 \includegraphics[width=\linewidth]{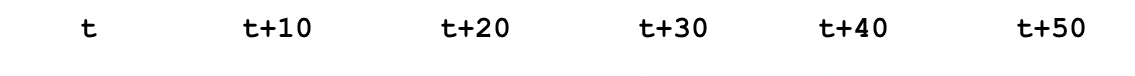}

 \caption[]{Effect of introducing damage to fully-formed geckos; columns are at $t=0, 10, 20, 30, 40, 50$ after damage 
 introduced. From top to bottom: no damage (reference case), front-right leg removed, front-left leg removed, both left 
 legs and part of body removed. All seeds are as top row (i.e.\ all legs). }

 \label{fig:gecko_damage}
\end{figure}

\begin{figure*}[t!] 
 \centering
 \subfloat[ Fully-grown heart organism responding to two ($t=0, 100$) external signals. \label{fig:results:external:heart} ] { \centering %
 \includegraphics[width=\linewidth]{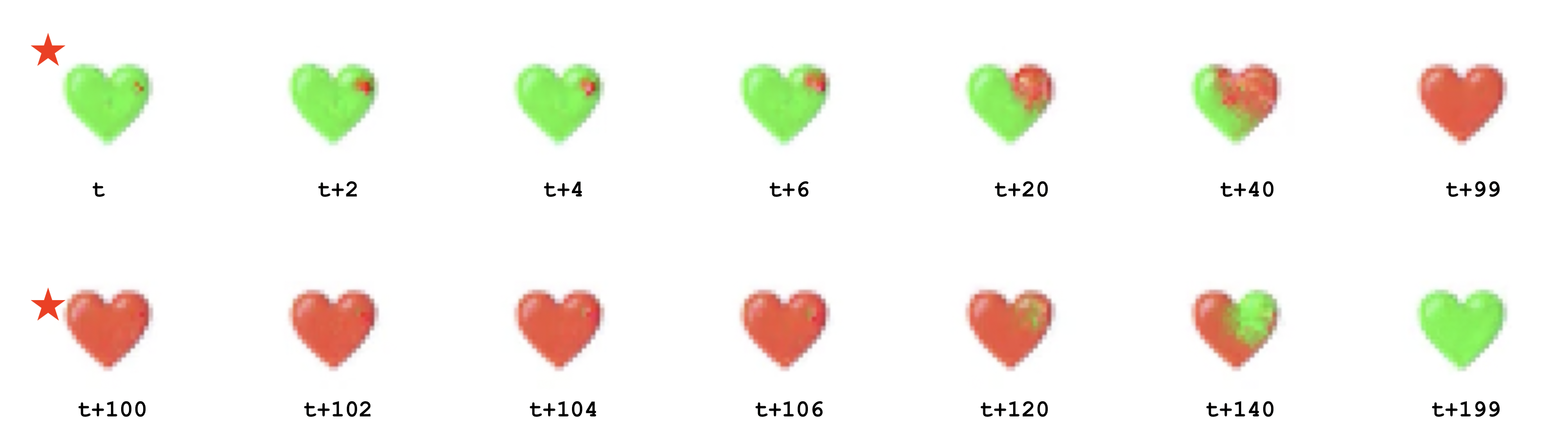} } \\
 \subfloat[ Fully-grown gecko organism responding to two ($t=0, 100$) external signals. \label{fig:results:external:gecko} ] { \centering %
 \includegraphics[width=\linewidth]{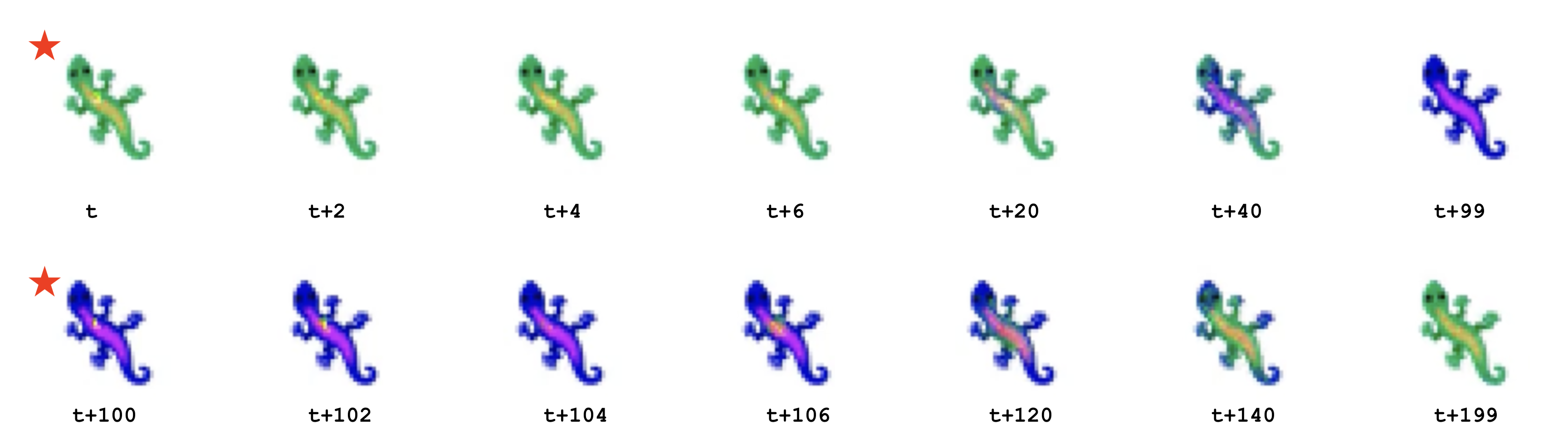} }

 \caption[]{Two-way colour change as a response to an external signal. The organism used is the fully-trained form of 
 the NCA which can change colour back and forth repeatedly. Column labels indicate number of timesteps after signal 
 provided. Organism on bottom row is continuation of top row. Red star indicates timestep with signal provided.} 
 \label{fig:results:external}
\end{figure*}

%\begin{figure*}[h!] 
% \centering
% \includegraphics[width=\linewidth]{gecko_colourchange_greenblue.png}
% \caption[]{colour change from external signal}
% \label{fig:results:external:gecko}
%\end{figure*}
% While the NCAs tend to be fragile and prone to disintegration, it 

To test whether the NCA relies on the current shape of the organism to encode which organism it is, we damaged the 
grown organisms and tested whether they could regenerate back to their original forms. Fig.~\ref{fig:gecko_damage} 
shows the outcome of this test. In this example, all the organisms were grown with a seed cell that encodes for all 
four legs. Upon damaging the organism, the damage is repaired within 40--50 timesteps, but often results in a 
different organism to which it started. This shows that---in this case---the NCA is not encoding information about the 
organism into the cells, it relies on the current form of the organism to encode the information instead.

This section has presented a range of different tests showing how NCAs can be trained to respond to internal 
(genomically-coded) signals. This provides a reasonable starting point for further investigation into the role of the 
seed cell for providing a genotype for an NCA-grown artificial organism.

\subsection{External Signals: Colour Changes}

While the previous tests were focussed on the ability of the NCA to respond to internal signals encoded in the seed 
cell of the organism, this section focusses on the ability of the NCA to respond to external signals provided to the 
fully-grown organism from the environment. For this, we will make use of the new read-only channel added to the NCA 
state vector (see fig.~\ref{fig:seed}), and will train the NCA to change colour when it receives a signal on this 
channel. The external signal will be provided to the fully-grown organism through one pixel for one timestep.

Fig.~\ref{fig:results:external} shows the result of training an NCA with this regime. In early training attempts, the 
NCA would learn to change from green to red, and back to green, but was unable to react to any further signals. This 
suggests that while the NCA appeared to be responding to the signal, it was actually replacing the organism with a 
different organism that was unable to respond to subsequent signals. This is not unlike the `adversarial takeover' 
NCAs of \citet{cavuoti_adversarialtakeoverneural}, but with a single neural network and a single pixel, rather than 
using multiple trained networks and multiple pixels.

To address this problem, we provide the NCA with the signal multiple times throughout the training process (e.g.\ if 
the organism is sampled from the training pool into the current training batch, we might provide another signal to 
it). The signal is still only presented for a single timestep, and we give sufficient time for the organism to change 
between signals. This resulted in much more stable behaviour, with the organism able to switch between the two colours 
when it receives the signal, seemingly as many times as it receives the signal.\footnote{informally, successfully 
tested up to 28 signals}

To confirm that this behaviour would work on more complicated organisms (i.e.\ with more than just one block colour), 
we also tested using gecko target images. As shown in fig.~\ref{fig:results:external:gecko}, this works just as well as 
the heart targets.

In order to ensure the NCA was not just growing a new organism over the top of the old organism, we introduced a small 
amount of `jitter' to the provided signal, i.e.\ we randomly moved the location of the signal within a small region. 
This jitter pushed the NCA to learn to {\em respond} to the signal. To ensure that the NCA had not just memorised all 
the possible locations at which the signal might be provided, we tested the response of the organism with signals 
outside the region used in training. While these generally took longer to respond, the NCA still responded in the 
expected manner. If the location was too far away from the originally-trained regions, the response did not occur, but 
this is not dissimilar to ensuring a signal is provided to the correct nerve to trigger a muscle contraction.

%---if you don't get it right, the muscle doesn't contract.

\section{Discussion}

In this paper we have demonstrated that NCAs can be trained to respond to signals. We have trained the NCA to respond 
to two types of signal: those encoded in the seed cell (internal) and those received from the environment (external). 
In both cases the signal is transient---it is only provided for a single timestep and through a single pixel, the main 
difference is whether the organism is fully-formed when the signal is received.

% morphogenesis / programmed cell death / cell division (apotosis / mitosis)

Given that NCAs were originally conceived as a model of morphogenesis, the inclusion of signal response into the model 
opens new avenues for research along these lines, especially focussed on programmed cell death (apoptosis) and cell 
division (mitosis). We have already started looking at two areas: innervation of muscles in the NCA model, and how we 
might be able to adapt the training mechanism presented here to develop mitotic behaviour, where a fully-grown cell 
receives a signal and splits into two cells.
 
Furthermore, this work contributes to the development of an NCA in which the neural network acts as the machinery for 
expressing genetic information stored in the seed cell~\citep{dreyfuss_messengerrnabinding}. This `generalised' NCA 
should then be able to produce the corresponding phenotype for any seed cell provided to it, and may allow us to study 
the computational mechanisms around gene expression in an artificial substrate. This could also be a reasonable point 
at which to consider training the NCA on real cellular data.  

% black box nature of neural networks, in particular in comparison to genetic algorithms?

One of the disadvantages of using NCAs to study biological phenomena is the lack of transparency in neural network 
models. This is a reasonable consideration that needs further work, but in this case we would argue that the lack of 
transparency can be considered a tradeoff with our ability to engineer the emergent behaviour of the system---a 
notoriously difficult task~\citep{polack_emergentpropertiesrefine, stepney_engineeringemergence}.

To conclude, we have presented an approach for training NCAs such that they are able to respond to internal and 
external signals. We presented results focussed on the ability of the NCA to grow into different organisms based on 
genomically-coded signals, and to change colour based on environmental signals. Overall these contribute to the 
development of NCAs as a model of artificial morphogenesis, and pave the way for future developments embedding dynamic 
behaviour into the NCA model.

%\clearpage{}

\footnotesize
%\bibliographystyle{apalike}
%\bibliography{paper}

\end{document}